\documentclass{article}
\usepackage{amsmath}
\usepackage{amssymb}
\usepackage{xcolor} 
\usepackage{enumitem}  
\usepackage{graphicx}
\usepackage[most]{tcolorbox} 
\definecolor{peach}{RGB}{255, 218, 185}      
\definecolor{sage}{RGB}{160, 200, 120}       
\definecolor{rose}{RGB}{255, 182, 193}       
\definecolor{mint}{RGB}{245, 255, 250}       
\definecolor{sky}{RGB}{176, 196, 222}        
\definecolor{lavender}{RGB}{230, 230, 250}   
\definecolor{coral}{RGB}{255, 127, 80}       

\definecolor{peachdark}{RGB}{255, 160, 120}  
\definecolor{sagedark}{RGB}{100, 160, 80}    
\definecolor{rosedark}{RGB}{255, 120, 150}   
\definecolor{mintdark}{RGB}{180, 220, 200}   
\definecolor{skydark}{RGB}{100, 150, 200}    
\definecolor{lavenderdark}{RGB}{180, 180, 220} 

\newcommand{\blockspacing}{0.2em}   

\newenvironment{GoalBlock}[1][]{%
  \begin{tcolorbox}[colback=blue!15, colframe=white, boxrule=0pt, left=0cm, right=0cm, top=0cm, bottom=0cm, before skip=\blockspacing, after skip=\blockspacing]%
  \raisebox{-0.5ex}{\tcbox[colback=blue!40, colframe=blue!40, boxrule=0pt, arc=3pt, left=0pt, right=0pt, top=0pt, bottom=0pt, nobeforeafter]{\small\textcolor{white}{Goal}}} %
  \small
}{%
  \end{tcolorbox}%
}

\newenvironment{RealDataBlock}[1][]{%
  \begin{tcolorbox}[colback=peach!20, colframe=white, boxrule=0pt, left=0cm, right=0cm, top=0cm, bottom=0cm, before skip=\blockspacing, after skip=\blockspacing]%
  \raisebox{-0.5ex}{\tcbox[colback=peachdark, colframe=peachdark, boxrule=0pt, arc=3pt, left=0pt, right=0pt, top=0pt, bottom=0pt, nobeforeafter]{\small\textcolor{white}{Real Data}}} %
  \small
}{%
  \end{tcolorbox}%
}

\newenvironment{LabelBlock}[1][]{%
  \begin{tcolorbox}[colback=sky!20, colframe=white, boxrule=0pt, left=0cm, right=0cm, top=0cm, bottom=0cm, before skip=\blockspacing, after skip=\blockspacing]%
  \raisebox{-0.5ex}{\tcbox[colback=skydark, colframe=skydark, boxrule=0pt, arc=3pt, left=0pt, right=0pt, top=0pt, bottom=0pt, nobeforeafter]{\small\textcolor{white}{Label}}} %
  \small
}{%
  \end{tcolorbox}%
}

\newenvironment{JustificationBlock}[1][]{%
  \begin{tcolorbox}[colback=sage!20, colframe=white, boxrule=0pt, left=0cm, right=0cm, top=0cm, bottom=0cm, before skip=\blockspacing, after skip=\blockspacing]%
  \raisebox{-0.5ex}{\tcbox[colback=sagedark, colframe=sagedark, boxrule=0pt, arc=3pt, left=0pt, right=0pt, top=0pt, bottom=0pt, nobeforeafter]{\small\textcolor{white}{Introspection}}} %
  \small
}{%
  \end{tcolorbox}%
}

\newenvironment{AlternateActionBlock}[1][]{%
  \begin{tcolorbox}[colback=rose!20, colframe=white, boxrule=0pt, left=0cm, right=0cm, top=0cm, bottom=0cm, before skip=\blockspacing, after skip=\blockspacing]%
  \raisebox{-0.5ex}{\tcbox[colback=rosedark, colframe=rosedark, boxrule=0pt, arc=3pt, left=0pt, right=0pt, top=0pt, bottom=0pt, nobeforeafter]{\small\textcolor{white}{Alternate Action}}} %
  \small
}{%
  \end{tcolorbox}%
}

\newenvironment{ProposedActionBlock}[1][]{%
  \begin{tcolorbox}[colback=coral!20, colframe=white, boxrule=0pt, left=0cm, right=0cm, top=0cm, bottom=0cm, before skip=\blockspacing, after skip=\blockspacing]%
  \raisebox{-0.5ex}{\tcbox[colback=coral, colframe=coral, boxrule=0pt, arc=3pt, left=0pt, right=0pt, top=0pt, bottom=0pt, nobeforeafter]{\small\textcolor{white}{Proposed Action}}} %
  \small
}{%
  \end{tcolorbox}%
}

\newcommand{\Scenes}{\ensuremath{\mathcal{X}}}
\newcommand{\Actions}{\ensuremath{\mathcal{A}}}
\newcommand{\Labels}{\ensuremath{\mathcal{Y}}}

\newcommand{\Nodes}{\ensuremath{\mathcal{V}}}

\newcommand{\Types}{\ensuremath{\mathcal{T}}}
\newcommand{\emb}{\ensuremath{\phi}}
\newcommand{\simfn}{\ensuremath{S}}
\newcommand{\UNSAFE}{\texttt{UNSAFE}}
\newcommand{\SAFE}{\texttt{SAFE}}
\newcommand{\REASONABLE}{\texttt{REASONABLE}}

\usepackage[preprint]{corl_2025} 

\title{The Case for Negative Data: From Crash Reports to Counterfactuals for Reasonable Driving}

\author{
\textbf{Jay Patrikar$^{1,}$\thanks{This work was done in-part as an intern at NVIDIA Research and a PhD student at the CMU.}~~, Apoorva Sharma$^{2}$, Sushant Veer$^{2}$, Boyi Li$^{2}$,} \\
\textbf{Sebastian Scherer$^{3}$, Marco Pavone$^{2,4}$} \\
$^1$Field AI \quad $^2$NVIDIA \quad $^3$Carnegie Mellon University \quad $^4$Stanford University \\
}

\begin{document}
\maketitle


\begin{abstract}
Learning-based autonomous driving systems are trained mostly on incident-free data, offering little guidance near safety–performance boundaries. Real crash reports contain precisely the contrastive evidence needed, but they are hard to use: narratives are unstructured, third-person, and poorly grounded to sensor views. We address these challenges by normalizing crash narratives to ego-centric language and converting both logs and crashes into a unified scene–action representation suitable for retrieval. At decision time, our system adjudicates proposed actions by retrieving relevant precedents from this unified index; an agentic counterfactual extension proposes plausible alternatives, retrieves for each, and reasons across outcomes before deciding. On a nuScenes benchmark, precedent retrieval substantially improves calibration, with recall on contextually preferred actions rising from 24\% to 53\%. The counterfactual variant preserves these gains while sharpening decisions near risk. 
\end{abstract}



\section{Introduction}
\begin{figure}[htb!]
    \centering
    \includegraphics[width=\linewidth]{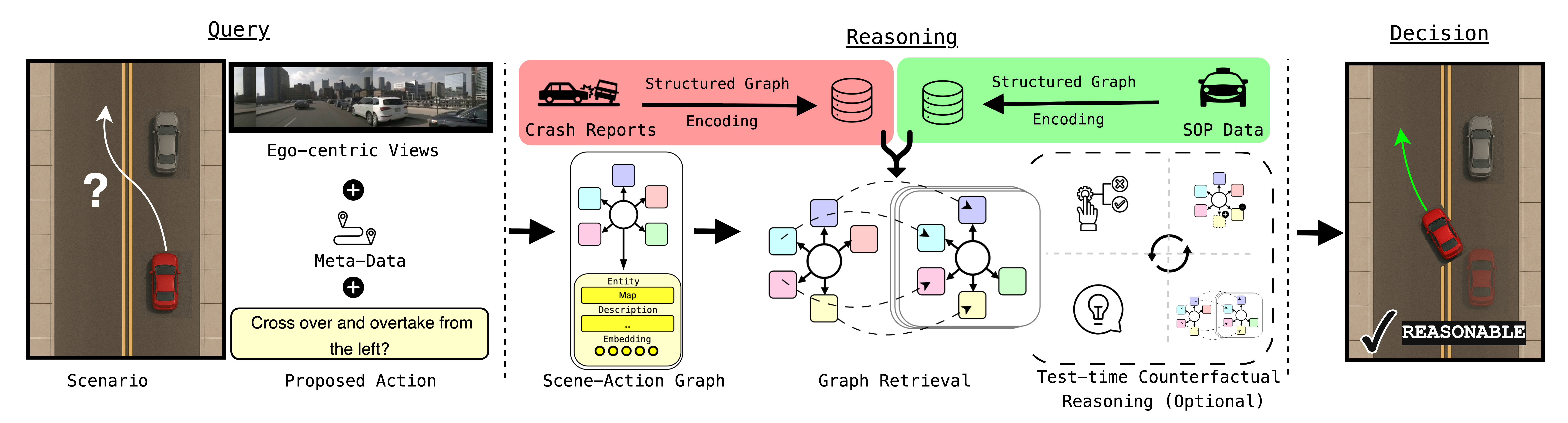}
    \caption{Overview of our pipeline. \textbf{Input}: a driving scene with front-facing video and a candidate ego action. \textbf{Processing}: convert the scene to a structured scene--action graph and retrieve precedents from a unified index of positive logs and crash narratives; optionally run an agentic counterfactual loop that proposes and evaluates alternative actions. \textbf{Output}: a justified label for the proposed action (\texttt{UNSAFE\,|\,SAFE\,|\,REASONABLE}) grounded in retrieved evidence.}
    \label{fig:motivation}
\end{figure}
End-to-end learning-based autonomous vehicle (AV) systems are trained primarily through imitation learning on positive, incident-free driving data \cite{caesar2020nuscenes, weng2024drive}. 
This data is typically collected by expert human drivers driving sensor-instrumented vehicles in a variety of driving scenarios, resulting in a dataset pairing the sensor observations that the AV will encounter with the action that the human driver chose in that moment.
While this data helps define ``good" driving that an AV should imitate, it does not provide direct supervision of what behaviors are to be avoided.
Some have aimed to address this gap through auxiliary reward functions defining a rules-based definition of risky driving \cite{patrikar2024rulefuser}, but such rules can be challenging to specify: Risk is difficult to quantify due to uncertainty over other road user's behaviors. Moreover, competent driving requires appropriately \textit{managing} the risk that is taken on to make progress; remaining stopped is the safest policy, but not competent driving behavior. 

Instead, in this work, we consider an alternative data-driven approach to provide negative supervision for AV decision making. Specifically, we explore the use of crash reports as a complementary source of driving knowledge. 
Agencies such as the National Highway Traffic Safety Administration (NHTSA) collect structured narrative accounts of real-world accidents, including the actions taken and the conditions under which failures occurred. While these reports lack the rich multimodal data of first-person human-driven AV logs, they contain valuable causal and contextual information that can support counterfactual reasoning. While these reports can't directly be used in policy training, recent advances in vision-language models (VLMs) capable of reasoning across sensor and text domains offer a compelling avenue for bringing such valuable sources of negative data into AV decision making.

In this paper, we study how negative data influences VLM reasoning capabilities in AV decision making tasks by developing a retrieval-augmented-generation (RAG) pipeline for AV safety adjudication. Specifically our contributions are as follows:
(i) a GraphRAG \citep{procko2024graph} style retrieval pipeline for both positive and negative driving precedent, using a unified structured language representation for both sensor-domain positive data and language-domain negative data; (ii) an agentic extension which uses additional test-time compute to reason about counterfactuals prior to making a safety judgment; (iii) evaluation of both approaches in terms of alignment with human judgement on the safety of possible actions in driving scenarios, showing the impact of negative crash report data on VLM decision making capability.

\section{Related Works}

\subsection{Safety Assessment of Motion Plans}
The increasingly black-box nature of autonomous driving systems raises significant challenges towards safety assessment of the motion plans they produce.  
Classically, methods drawn from reachability theory \cite{bansal2017hamilton} have been used to provide an understanding of ``criticality" of a particular interaction. Vanilla reachability uses the worst-case assumption on the behavior of other agents which makes it overly conservative limiting its use in real-world situations. While methods that assume a more reasonable behavior from the other agent have been proposed \cite{chakraborty2025safety,nakamura2023online}, the effectiveness of these methods in distribution shifts and high-dimensional data are not well studied. Uncertainty quantification (UQ) \cite{patrikar2024rulefuser,itkina2023interpretable} is another paradigm that can be used to identify unsafe scenarios given a dataset of safe scenarios. While UQ methods are effective in identifying scenarios that are far from the safe distribution, these methods fail to offer counterfactuals which are often essential to reason about safety \cite{zanardi2023counterfactual,roy2025fractional}. Rules \cite{stoler2024safeshift} and catalog-based \cite{weber2019framework} methods that rely on humans to craft hand-engineered rules have also been proposed, but these methods are painstaking to develop and can struggle with far edge-case scenarios that were overlooked during development. Yet another category of approaches, that fall within the sub-field of explainable AI, furnish human-understandable ``reasoning" that the model plausibly underwent for generating a particular outcome \cite{sado2023explainable}. Some popular methods include the use of visual saliency maps \cite{renz2022plant} and vision and language captions for providing decision explanations \cite{jin2023adapt,ben2022driving,yuan2024rag,xu2024drivegpt4} among others. However, these approaches predominantly focus on positive behaviors, limiting their ability to distinguish good from bad driving behaviors, unlike this paper.

\subsection{Leveraging Crash Report Data}
A significant bottleneck to using negative or crash data is its lack of availability. Most regular driving logs---e.g., nuScenes \cite{caesar2020nuscenes}, nuPlan \cite{nuplan}, and Waymo Open Motion Dataset \cite{ettinger2021large}---which include vectorized scene information only provide safe driving data. To address this limitation, NHTSA crash reports \cite{NHTSA_Crash_Viewer}, which are in the form of textual descriptions of real-world crashes, have been leveraged for synthesizing crash simulations \cite{tan2023language,li2025crashagent,wang2025enhancing}. However, it remains unclear how realistic these simulations are to be able to stake a safety claim using them. Directly using the crash reports without converting them to simulations first can alleviate the need for validating the simulations. However, most prior works that directly use the crash reports look at aggregate bechmarking of the safety of autonomous driving rather than a case-by-case safety analysis \cite{scanlon2024benchmarks}. In this paper, we develop an approach that uses RAG to retrieve the nearest scenario to the current one from a database that consists of the NHTSA crash reports along with regular driving data from nuScenes and supply it to a VLM-based safety monitor. Importantly, we show that the use of the crash reports help reduce the over-conservatism inherent to VLM-based monitors.

\section{Problem Statement}
To investigate the impact of negative data on VLM decision-making, we create a benchmark based on action adjudication.
Specifically, we set up a retrieval-augmented pipeline for a multi-class classification problem.
Given a driving scene $x \in \Scenes$, and a candidate action $a \in \Actions$, the VLM is asked to classify whether action $a$ is either \UNSAFE{} (would likely result in a collision, rule violation, or hazardous interaction), \SAFE{} (legal and physically safe, but not an action that a reasonable driver would choose), and \REASONABLE{} (safe, and likely to be chosen by a reasonable driver). 
We distinguish \SAFE{} and \REASONABLE{} to better test the VLM's ability to adjudicate safety in alignment with human drivers -- often, the \REASONABLE{} action is the one that takes on the appropriate level of risk to maintain reasonable progress; marking \REASONABLE{} actions as \UNSAFE{} is an indication of an overconservative safety assessment.

We represent the scene $x$ in a VLM compatible manner using (i) a forward-facing camera image and (ii) a natural-language description of the scene. We choose a discrete action space containing the following high-level actions: 
\[
\Actions = \left\{ \begin{array}{c} \texttt{MERGE LEFT, TURN LEFT, NUDGE LEFT,} \\ \texttt{STRAIGHT, STOP, ACCELERATE, DECELERATE,} \\ \texttt{NUDGE RIGHT, TURN RIGHT, MERGE RIGHT} \end{array} \right\}
\]

For a given scene-action pair $(x,a)$ we query the VLM to predict the label $y \in \Labels = \{\UNSAFE{},\SAFE{},\REASONABLE{}\}$, and use the classifier performance as a signal for VLM safety adjudication capabilities. 

\section{A Unified Data Representation for Structured Retrieval of Positive and Negative Data}
\label{sec:unified-repr}
\begin{figure}[t]
    \centering
    \includegraphics[width=\linewidth]{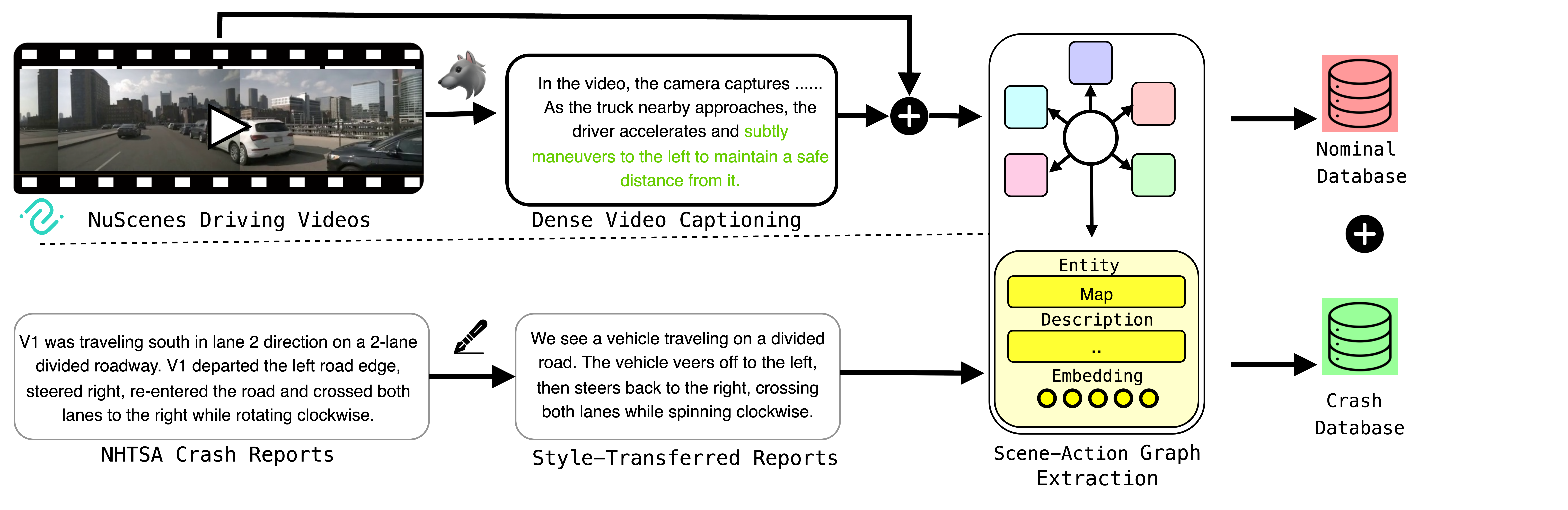}
    \caption{Unified representation. Positive driving logs and negative crash narratives are converted to structured scene–action graphs with canonical node types. Nodes carry natural language summaries and embeddings for GraphRAG retrieval.}
    \label{fig:unified-repr}
\end{figure}

A key challenge in enabling a decision maker to reason about both positive and negative driving examples is unifying the starkly different types of data -- high-quality, annotated camera data in the case of positive driving data, and relatively terse, technical crash reports in the case of negative data. 
We construct a unified retrieval corpus by converting both crash reports and AV driving logs into a shared structured representation suitable for retrieval.

In particular, we embed each into structured \textit{scene-action graphs}. Each graph $G$ consists of a set of nodes, each with a type $t \in \Types$ where the set of types is $\Types = \{ \mathsf{Ego}, \mathsf{Obstacles}, \mathsf{Map} \}$. Each node has an associated natural language description that summarize the state or behavior of each element in the scene. 
For positive data, we construct this graph using a pipeline involving video summarization using a VLM, followed by LLM processing of the result into the structured representation. For negative crash report data, we use an LLM-only pipeline to obtain the graphs. For details on these pipelines, and examples of inputs the output structured representation, see Appendix~\ref{app:prompts} and Figure~\ref{fig:unified-repr}.

These graphs serve as the corpus for GraphRAG-style retrieval: the text of each node is converted to a vector via NV-EmbedQA-Mistral-7B-v2, a state-of-the-art text embedding model optimized for retrieval tasks
\(\emb\). For a query graph \(G_q\) and a corpus graph \(G\), we compute similarity as a sum over node types, where the graph similarity for a node type is highest pairwise similarity between nodes of the same type:
\begin{align*}
    \simfn(G_q,G) 
    &= \sum_{t\in\Types} w_t\, \max_{v\in\Nodes_t(G_q)}\, \max_{u\in\Nodes_t(G)} 
        \; \mathrm{Sim}_\mathrm{cos}\big(\emb(d(v)),\emb(d(u))\big) \\
    &\text{where } \Nodes_t(G)=\{u\in\Nodes(G) \,:\, \tau(u)=t\},\; w_t\ge 0,\; \sum_t w_t=1.
\end{align*}
where $\Nodes(G)$ is the set of nodes of graph $G$, and $\tau(u)$ returns the type of node $u$, and $\mathrm{Sim}_\mathrm{cosine}$ is the cosine similarity of two vectors.
We use top-\(k\) retrieval under \(\simfn\) to form the precedent set for decoding.
By explicitly comparing the query scene against the corpus on each relevant axis, such graph-based retrieval enables more precise and controllable action-conditioned retrieval than directly using holistic embeddings of the full scene descriptions.

\subsection{Structuring Negative Data from Crash Reports}
We process NHTSA crash reports through a three-step pipeline: (1) sourcing free-text crash narratives, (2) style normalization to convert third-person templated language into ego-centric descriptions, and (3) scene-action graph construction using prompt-based extraction to identify canonical entities (\textit{Map, Ego, Ego Action, Obstacles, Obstacles Action, Collision/No Collision}). Each node is embedded to enable fine-grained graph-level similarity for retrieval. See Appendix~\ref{app:data-structuring} for detailed implementation.

\subsection{Structuring Positive Data from Driving Logs}
We extract positive driving scenarios from the nuScenes dataset and process them through a three-step pipeline: (1) sourcing sensor-rich logs, (2) dense video captioning to summarize driving segments, and (3) scene-action graph construction using prompt-based extraction to identify canonical entities (\textit{Ego, Ego Action, Obstacles, Map}). Each node is embedded to support fine-grained GraphRAG retrieval. See Appendix~\ref{app:data-structuring} for detailed implementation.
  

\section{VLM Reasoning Engines for Adjudicating Driving Actions}
\begin{figure}[t]
    \centering
    \includegraphics[width=\linewidth]{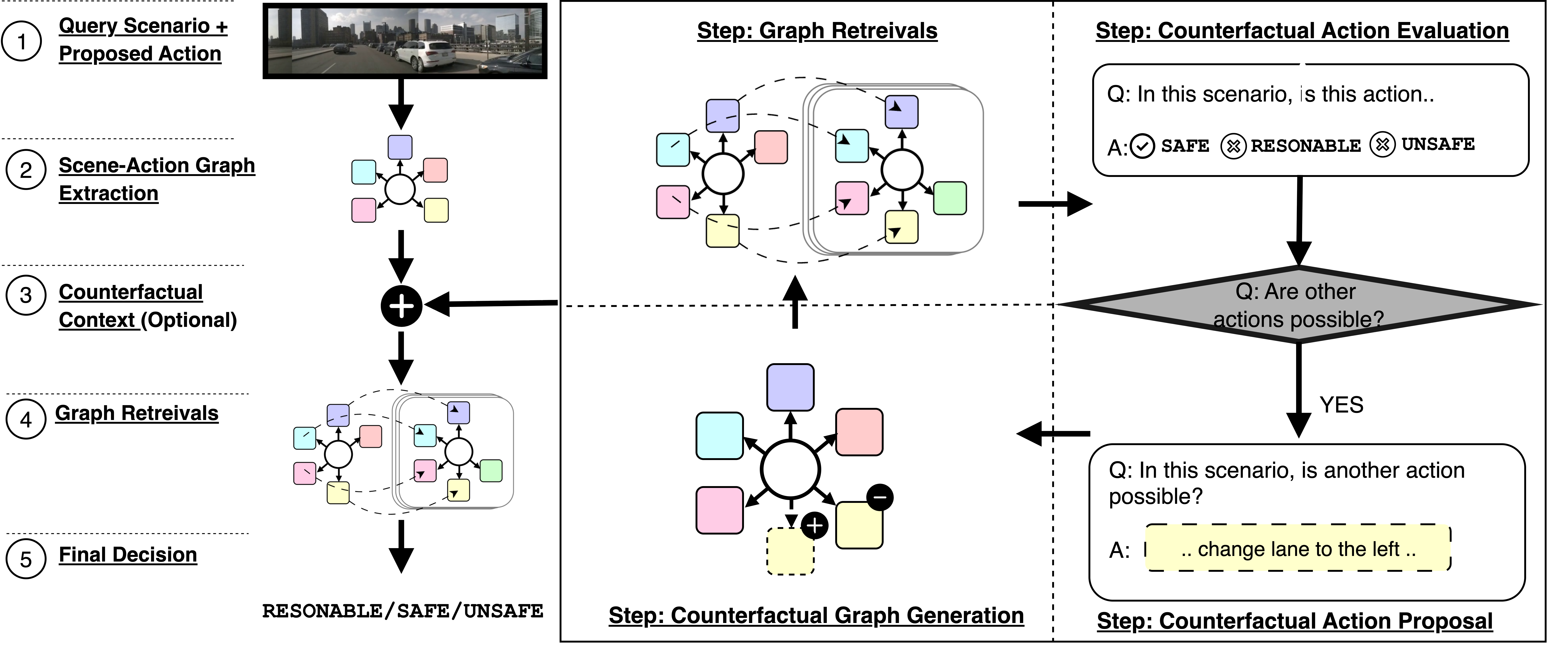}
    \caption{Reasoning engines. We compare a VLM-only baseline (only steps 1 and 5, no additional context), a one-step CoT with precedent retrieval via GraphRAG (1,2,4,5), and an agentic counterfactual loop that proposes alternatives and retrieves precedents for each before adjudication (1,2,3,4,5).}
    \label{fig:reasoning-engines}
\end{figure}
In order to study the impact of negative data on adjudicating the safety of driving actions, we compare three reasoning paradigms that incorporate precedent data to varying degrees: (1) a base vision-language model (VLM) operating without external context \(f_{\text{base}}\), (2) a VLM augmented with precedent retrieval via GraphRAG and chain-of-thought (CoT) decoding \(f_{\text{rag}}\), and (3) an agentic extension that iteratively explores counterfactuals to improve robustness \(f_{\text{agentic}}\), each illustrated in Figure~\ref{fig:reasoning-engines}.

For all reasoning engines, we use GPT-4o as the primary reasoning engine, which provides strong multimodal reasoning capabilities for both visual scene understanding and textual precedent analysis. We leave the impact of the choice of different VLMs and spatially-aware VLMs, including systems that use specialized models for spatial reasoning like GroundingDINO \cite{liu2023groundingdino}, for future conference-level iterations of this work.

Next, to study the impact of providing driving precedent, we provide the VLM with context retrieved using the GraphRAG approach detailed above, and augment the prompt to encourage chain-of-thought (CoT) decoding to factor in the retrievals into the judgment, $f_\mathrm{RAG}$. Finally, we explore the impact of additional test-time compute by constructing an agentic extension that iteratively explores counterfactuals, performing multiple rounds of retrievals from the precedent corpus, $f_\mathrm{agentic}$. We detail each of these approaches below. 


\subsection{VLM-Only Baseline}
In the simplest configuration, the VLM receives a natural language description of the driving scene and a proposed alternate action for the ego vehicle. It is prompted to reason about the action’s legality, safety, and efficiency, and assign one of three outcome labels: \UNSAFE, \SAFE, or \REASONABLE. This method reflects a purely generative model without access to precedent scenarios, which we use as a baseline for comparison.

\subsection{One Step RAG + Chain-of-Thought Decoding}
To enrich the VLM’s judgment with external behavioral context, we augment it with precedent retrieval via a GraphRAG-style method tailored to our structured graph format. 

Given a query scene and proposed ego action, we construct the query graph $G_q$, setting the \textit{Ego Action} node with the candidate, and retrieve the top-$k$ most similar graphs from the precedent corpus according to the similarity function $\simfn(G_q, G)$ detailed in Section~\ref{sec:unified-repr}. 

We append the full textual description of the retrieved scenes, as well as their class (positive/negative) to the VLM context, together with a CoT prompt which encourages the VLM to draw analogies to the retrieved scenarios when adjudicating the action. See the Appendix for more details and the prompt template used.
\subsection{Agentic Counterfactual GraphRAG}

The agentic counterfactual loop builds on recent trends in test-time reasoning and planning, where models dynamically adjust their computational effort based on task complexity. 
%
This aligns closely with recent Agentic paradigms like ReAct \cite{yao2023react}, Reflexion \cite{shinn2023reflexion}, and test-time counterfactual planning in embodied agents \cite{sun2025llapa}, where additional reasoning steps, retrievals, or internal simulations are performed only when needed. 
Rather than producing a single output in one pass, these systems reframe decision-making as a multi-step process—evaluating, reflecting, and refining based on intermediate outputs.

In our setting, each retrieval–evaluation loop represents a unit of deliberation. Simple scenarios —e.g., clear STOP or obvious collision—may resolve in one step. But complex scenarios may require multiple iterations of proposal, evidence gathering, and adjudication before a confident classification is made. To enable this, we allow the VLM to propose alternative actions with which to query the current scene. Instead of issuing a single judgment, the model actively generates and evaluates plausible alternatives, each grounded in precedent cases. This mirrors human-like counterfactual reasoning: “What if I turned instead of braking? Have others made similar choices in this scenario—and what happened next?” The model’s internal state evolves over time, accumulating both counterfactual actions and their justifications to make better-informed final decisions.


The process follows a generation–retrieval–evaluation loop, implemented as a LangGraph \cite{LangChainDocumentation} pipeline:

\begin{itemize}
\item \textbf{(Step 1) Generation:} A new counterfactual action is proposed. The model internally reflects on prior retrievals and decisions to avoid repetition and focus on unexplored decision boundaries.
\item \textbf{(Step 2) Retrieval:} The query graph is updated with the proposed action, and top-$k$ similar scenes are retrieved from the precedent database using GraphRAG. These retrievals enhance behavioral diversity while preserving semantic alignment. 
\item \textbf{(Step 3) Evaluation:} Retrieved scene descriptions are used to assess the plausibility and safety of the counterfactual action. Evaluation is guided by precedent-aligned reasoning and incorporates justification feedback, which is used to refine future proposals.
\end{itemize}

This cycle repeats till the agent decides to stop or for a max number of iterations. At each step, introspective reasoning—akin to ReACT or Reflexion—allows the model to learn from previous steps and improve future iterations. The accumulated evidence—drawn from diverse plausible behaviors—forms a more robust context for evaluating the original action. The final classification for the candidate action is made after this deliberative process, incorporating precedent-informed reasoning comparing the candidate action against a richer behavioral space.


\section{Results and Discussions}

We investigate how structured precedent data—particularly crash reports—can inform the classification of counterfactual driving actions as UNSAFE, SAFE, or REASONABLE. Human annotators labeled a balanced dataset of 1,275 action-scene pairs, based on visual scenes, textual descriptions, and ten standard driving actions. We evaluate how vision-language models (VLMs) perform across four configurations: without retrieval (VLM-only), with precedent retrieval (VLM+RAG), with RAG using only positive examples (VLM+RAG-PosOnly), and with an agentic counterfactual refinement mechanism (Agentic RAG). This section summarizes our key findings and hypotheses, see the Appendix for more details on experimental set-up.

\begin{figure}[t]
    \centering
    \includegraphics[width=\linewidth]{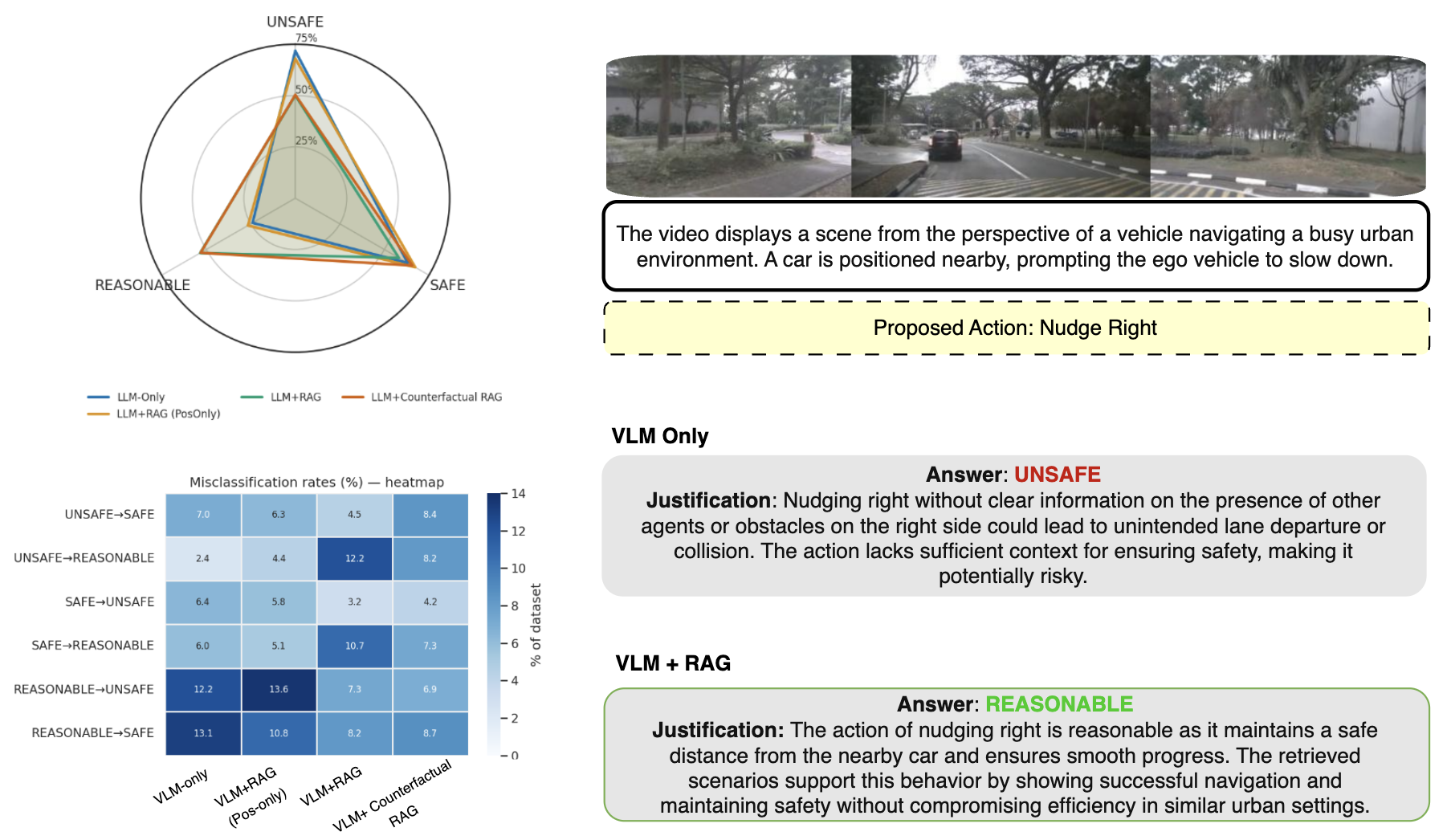}
    \caption{Overall results. Left: consolidated quantitative summary combining class-wise recall and misclassification breakdown across four engines (VLM-only, VLM+RAG (PosOnly), VLM+RAG, Agentic RAG). Right: qualitative example where the ground-truth outcome is \REASONABLE{}. The VLM-only model labels the proposed action \UNSAFE{}, while precedent-augmented VLM+RAG correctly judges it \REASONABLE{}, citing retrieved precedents. See text for discussion of recall gains and critical error reductions.}
    \label{fig:results-summary}
\end{figure}

\paragraph{Retrieval Enables Nuanced Reasoning Beyond Binary Safety Judgments}
Retrieval-augmented models significantly improve the ability to distinguish between actions that are merely safe and those that are both safe and contextually preferred. Without access to precedents, the VLM-only model overpredicts danger—labeling 37\% of \REASONABLE{} actions as \UNSAFE{}. By incorporating retrieved scenarios, the VLM+RAG model increases \REASONABLE{} recall from 24\% to 53\%, demonstrating a more refined understanding of assertive yet appropriate behavior. This improvement, however, comes with a tradeoff: \UNSAFE{} recall drops from 72\% to 50\%, suggesting 
the base VLM tends towards conservative caution without precedent, but when offered the possibility of negative precedent but lacking a similar enough negative example, may falsely assume an action is \SAFE{}. In practice, we observe that many \UNSAFE{}\,$\to$\,\SAFE{} errors arise when retrieved reports are not semantically close to the scene at hand. Because our crash corpus is limited in size and coverage (we do not use the full NHTSA dataset due to budget constraints), the system can implicitly treat the absence of a matching negative precedent as weak evidence of safety. We expect this effect to diminish with a larger and more diverse set of crash reports. As an additional mitigation, when no sufficiently similar precedent is retrieved, the model could be prompted to explicitly reason about potential failure modes rather than defaulting to \SAFE{}. 
These shifts suggest that precedent retrieval largely improves the model's ability to align with human preferences on the actions that human labellers deem efficient -- those that appropriately manage risk while maintaining efficiency.

\paragraph{Negative Precedents Are Crucial for Contrastive Reasoning}
 When crash reports are withheld from the retrieval pool, \REASONABLE{} classification drops sharply---from 53\% to 27\%---highlighting the model’s inability to recognize assertive but acceptable behavior without negative examples to identify the threshold at which actions become unacceptable. At the same time, \UNSAFE{} recall improves from 50\% to 68\%, but with 41\% of \REASONABLE{} actions misclassified as \UNSAFE{}---worse than the VLM---only baseline. This indicates that in the absence of negative examples, the VLM leans towards a brittle, overly cautious prior. Without real-world accounts of failure, the model defaults to treating any deviation from passive behavior as unsafe. Negative precedents restore balance by providing grounded evidence for when safety is truly violated, leading to more selective and meaningful \UNSAFE{} predictions.

\paragraph{Agentic Counterfactual Adjudication Enhances Decision Boundary Calibration}
Compared to vanilla VLM+RAG, the agentic counterfactual approach maintains strong \REASONABLE{} recall (53\%) while improving both precision and calibration at the class boundaries. SAFE recall increases from 58\% to 66\%, indicating better recognition of legally sound but less aggressive decisions. \UNSAFE{}-to-\REASONABLE{} errors drop notably from 36\% to 24\%—a sign of improved caution on borderline dangerous actions. These improvements suggest that the generate–retrieve–evaluate cycle helps the model simulate plausible alternatives, weigh them with precedent, and return more robust, context-aware classifications than flat retrieval alone.

\section{Limitations}
\label{sec:limitations}

While our results are encouraging, several limitations remain, and we view this work as a workshop-stage exploration rather than a production-ready system.

\textbf{Spatial reasoning in VLMs.} Current vision–language models struggle with fine-grained spatial reasoning (e.g., occlusions, relative lane geometry, and precise right-of-way logic). Ongoing efforts aim to improve grounding and geometric consistency via better visual grounding, structured scene representations, and spatially-aware prompting/fine-tuning \cite{liu2023groundingdino, chen2024spatialvlm}. Our approach partially mitigates this by retrieving precedent narratives that encode spatial context in natural language, but residual failures in spatial inference persist. We expect improvements in spatial reasoning capabilities will mitigate these in the future.

\textbf{Dataset and scope.} Our evaluation dataset is limited in size and coverage. The crash corpus used for negative precedents comprises only 1.4k reports, and our benchmark subsamples to 1,275 action–scene pairs (425 per class). Larger and more diverse datasets—both in nominal driving and in negative incidents—are needed to stress-test generalization, rare interactions, and long-tail scenarios beyond the present scope.

\textbf{Runtime and deployment.} The full pipeline may be too slow for strict real-time operation with current VLMs. That said, inference times are trending down, and our method is immediately useful for offline scenario mining and policy auditing. For online use, the graph index can be made more efficient through compression and hierarchical organization (e.g., coarse-to-fine node pruning), approximate nearest-neighbor search over node embeddings, and lightweight re-ranking. These changes would preserve precedent alignment while improving retrieval latency.
\section{Conclusion}
\label{sec:conclusion}

We introduced precedent-guided adjudication for autonomous driving by converting logs and crashes into a unified scene–action representation and retrieving precedents with VLM+RAG. We further proposed an agentic counterfactual VLM+RAG engine that proposes alternatives, retrieves precedents for each, and adjudicates across outcomes.

On a balanced nuScenes benchmark, VLM+RAG substantially improves calibration over a VLM-only baseline (e.g., recall on contextually preferred actions 24\%\,$\to$\,53\%), and the agentic counterfactual variant maintains these gains while further sharpening decisions near risk. Beyond AVs, precedent-guided, counterfactual adjudication can support domains that balance safety and performance under uncertainty. Future work includes scaling multimodal inputs, learning retrieval weights end-to-end, and integrating causal structure from simulators and maps.


\acknowledgments{This work was completed as part of NVIDIA Research Internship. This work also used Bridges-2 at Pittsburgh Supercomputing Center through allocation cis220039p from the Advanced Cyberinfrastructure Coordination Ecosystem: Services \& Support (ACCESS) program, which is supported by National Science Foundation grants \#2138259, \#2138286, \#2138307, \#2137603, and \#2138296. }


\bibliography{example}  
\clearpage
\appendix

\section{Data Structuring Details}
\label{app:data-structuring}

\subsection{Structuring Negative Data from Crash Reports}
\begin{description}[style=nextline,leftmargin=0em,labelsep=1em]
  \item[Step 1: Sourcing] We use free-text crash narratives from the National Highway Traffic Safety Administration (NHTSA) crash database. These narratives describe the sequence of events leading up to and following vehicle collisions, often written in a templated but unstructured format.

  \item[Step 2: Style Normalization via Prompt-Based Conversion] NHTSA crash reports contain detailed descriptions of real-world collisions, but are written in templated, third-person language with identifiers like "V1" and "V2" and cardinal directions (e.g., "eastbound"). To make them compatible with AV data and downstream retrieval, we apply a prompt-based style transfer procedure using a language model.

The model is prompted to reframe each scenario from the ego vehicle's perspective, replacing entity tags with relational terms (e.g., "ego vehicle," "car in front") and converting absolute references into relative spatial language. It also removes irrelevant technical detail while preserving causal structure and intent. This conversion is guided by a system prompt and a set of in-context examples (provided in the appendix). This step preserves causal structure while eliminating technical jargon and irrelevant formatting.

  \item[Step 3: Scene-Action Graph Construction]
Each normalized crash description is parsed into a structured \textit{scene-action graph}. We use a constrained prompt-based extraction method to identify entities from a fixed schema of node types:
\[
Nodes = \left\{ \begin{array}{c} \texttt{Map, Ego, Ego Action,} \\ \texttt{Obstacles, Obstacles Action,} \\ \texttt{Collision or No Collision} \end{array} \right\}
\]

For each node, we extract a canonicalized name, type, and natural language description. If multiple nodes of the same type exist (e.g., several nearby vehicles), their descriptions are merged into a unified summary under the same node type. This ensures compact yet complete semantic representations.

In addition to storing the textual descriptions, we compute a vector embedding for each individual node using NV-EmbedQA-Mistral-7B-v2, a pre-trained embedding model optimized for retrieval tasks. This allows us to retrieve scenes based on fine-grained graph-level similarity by computing similarity over sets of node embeddings, rather than relying on holistic sentence-level encodings alone.
\end{description}

\subsection{Structuring Positive Data from Driving Logs}
\begin{description}[style=nextline,leftmargin=0em,labelsep=1em]
  \item[Step 1: Sourcing] To complement crash reports, we extract positive driving scenarios from the nuScenes dataset, which provides sensor-rich logs of urban driving. Each scene includes ego vehicle pose, high-resolution camera footage, annotated tracks of surrounding agents, and detailed map context. However, this data is not naturally in a textual or structured narrative format and thus cannot be directly compared to processed crash reports.
  
  \item[Step 2: Dense Video Captioning] We begin by summarizing short temporal segments of driving using WOLF \cite{li2024wolf}, a vision-to-language model designed to produce natural language descriptions from multi-camera input and associated metadata. These captions describe what the ego vehicle did, its context (e.g., intersections, obstacles), and agent interactions. We additionally append structured metadata including agent types, their positions relative to the ego vehicle, and coarse map information (e.g., whether the ego is approaching a merger or intersection).
  
  \item[Step 3: Scene-Action Graph Construction] Each datapoint is converted into a structured \textit{scene-action graph} using a prompt-based extraction process, consistent with the format used for crash data. The model identifies canonical entities (e.g., \textit{Ego, Ego Action, Obstacles, Map}) along with their attributes and relationships. To ensure spatial grounding, the forward-facing ego camera image is included as part of the prompt context, anchoring the language model's interpretation to the actual visual layout of the scene. If multiple entities of a given type are present (e.g., several vehicles or pedestrians), their descriptions are merged into a single node. Each node is associated with both a natural language summary and a learned embedding, computed independently using NV-EmbedQA-Mistral-7B-v2, to support fine-grained GraphRAG retrieval.
\end{description} 
\vspace{-1.5cm}
\section{Prompt Templates and Examples}
\label{app:prompts}

This section demonstrates the conversation structure and prompt templates used in our proposed system. The following shows a complete multi-turn conversation illustrating how the system analyzes driving scenarios and provides safety assessments.

\begin{GoalBlock}
You are the decision-making brain of an autonomous vehicle engaged in a multi-step reasoning process. Your task involves analyzing scenarios, actions, and their alternatives through the following structured approach:

\vspace{0.5em}
\textbf{1. Initial Action Evaluation:}
\begin{itemize}[leftmargin=1.5em, itemsep=0.2em]
  \item Analyze the provided action in the current scenario
  \item Consider the retrieved similar scenarios to inform your judgment
  \item Assess safety (collision avoidance, traffic rule compliance) and performance (efficiency, comfort, goal achievement)
\end{itemize}

\vspace{0.5em}
\textbf{2. Counterfactual Generation \& Analysis:}
\begin{itemize}[leftmargin=1.5em, itemsep=0.2em]
  \item Generate physically possible alternative actions that could have been taken
  \item For each alternative:
    \begin{itemize}[leftmargin=1em, itemsep=0.1em]
      \item Use retrieved scenarios to evaluate its viability
      \item Consider both conventional and creative solutions
      \item Assess safety and performance implications
    \end{itemize}
  \item Repeat this process 2--3 times to explore different possibilities
\end{itemize}

\vspace{0.5em}
\textbf{3. Comparative Analysis:}
\begin{itemize}[leftmargin=1.5em, itemsep=0.2em]
  \item Compare all considered actions (original and alternatives)
  \item Use retrieved scenarios to support comparisons
  \item Consider trade-offs between safety and performance
\end{itemize}

\vspace{0.5em}
\textbf{4. Final Action Assessment:}
\begin{itemize}[leftmargin=1.5em, itemsep=0.2em]
  \item Evaluate any new proposed action against all previous insights
  \item Use the accumulated knowledge from retrievals and counterfactuals
  \item Make a final determination based on comprehensive analysis
\end{itemize}

\vspace{0.8em}
\noindent\textbf{A preferred action must:}
\begin{itemize}[leftmargin=1.5em, itemsep=0.2em]
  \item \textbf{Ensure safety} (collision avoidance, adherence to traffic laws)
  \item \textbf{Enhance performance} (efficiency, comfort, goal achievement)
\end{itemize}

\vspace{0.5em}
\noindent\textbf{Remember:}
\begin{itemize}[leftmargin=1.5em, itemsep=0.2em]
  \item Use retrieved scenarios actively in your reasoning
  \item Consider both conventional and creative alternatives
  \item Build upon insights from each iteration
\end{itemize}

\vspace{0.8em}
\noindent\textbf{Examples:}
$\{\verb+in_context_example_1+\}$

\vspace{0.5em}
\noindent Now begin your multi-step analysis of the scenario.
\end{GoalBlock}

\begin{RealDataBlock}
This video provides a first-person perspective of a vehicle navigating through an urban environment. Several cars are seen close to the ego vehicle throughout the drive. The ego vehicle adjusts its speed and maintains a safe distance from nearby cars, ensuring a smooth and safe progression through traffic.

\centering
\includegraphics[width=0.8\textwidth]{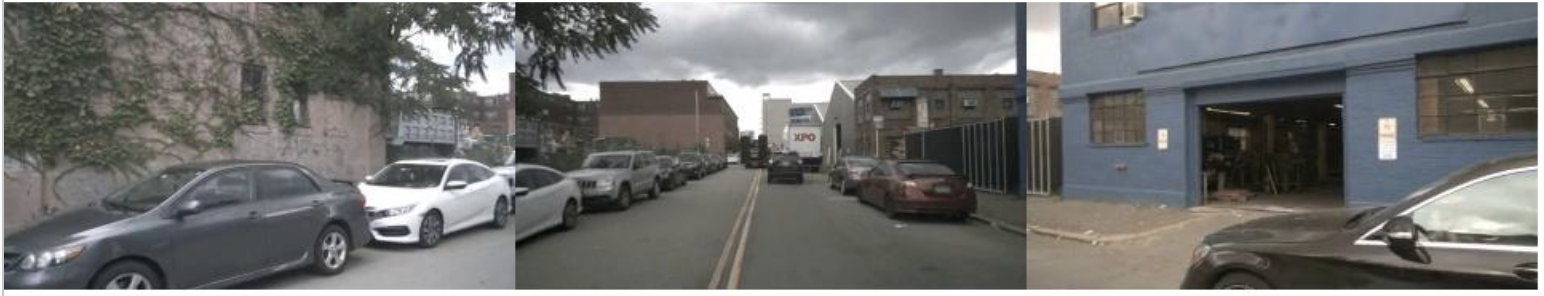}

Following is the meta-data:
The types of agents in proximity are: ['VEHICLE' 'VEHICLE' 'VEHICLE' 'VEHICLE' 'VEHICLE']
The relative positions of these agents with respect to ego agent are: ['REAR\_RIGHT' 'FRONT\_RIGHT' 'LEFT' 'REAR\_LEFT' 'FRONT\_LEFT']
The ego vehicle is on a road with no mergers or intersections.
Retrieved scenario:
\end{RealDataBlock}

\begin{JustificationBlock}
A feasible alternate action is to engage in a temporary stop to assess the surroundings and re-evaluate the situation. [Detailed reasoning omitted...]
\end{JustificationBlock}

\begin{AlternateActionBlock}
temporary stop to assess surroundings
\end{AlternateActionBlock}

\begin{JustificationBlock}
The action of performing a temporary stop to assess surroundings is classified as SAFE. \textcolor{gray}{[Detailed analysis omitted...]}
\end{JustificationBlock}

\begin{LabelBlock}
SAFE
\end{LabelBlock}

\begin{JustificationBlock}
A feasible alternate action is to employ a brief deceleration followed by a gentle arc to the left within the current lane. [Detailed reasoning omitted...]
\end{JustificationBlock}

\begin{AlternateActionBlock}
brief deceleration followed by a gentle arc to the left within the lane
\end{AlternateActionBlock}

\begin{JustificationBlock}
The action of brief deceleration followed by a gentle arc to the left within the lane is classified as REASONABLE. \textcolor{gray}{[Detailed analysis omitted...]}
\end{JustificationBlock}

\begin{LabelBlock}
REASONABLE
\end{LabelBlock}

\begin{ProposedActionBlock}
NUDGE LEFT
\end{ProposedActionBlock} 

\begin{JustificationBlock}
... The retrieved scenarios indicate that slight lane adjustments, such as veering to the left, are effective for maintaining safe distances from other vehicles. This action ensures safety by avoiding close proximity to nearby cars and enhances performance by maintaining a smooth flow through traffic. \textcolor{gray}{[Detailed analysis omitted...]}
\end{JustificationBlock}

\begin{LabelBlock}
REASONABLE
\end{LabelBlock}
End of conversation.
\newpage
\section{Human Labelling Interface}
\label{app:labeller}

We developed a custom labelling interface to collect human annotations for our driving action classification task. The interface presents human labellers with a driving scene image and a text summary of the scenario, then asks them to evaluate each of the 10 standard driving actions from our problem statement as either "SAFE", "UNSAFE", or "REASONABLE".

\begin{figure}[h]
    \centering
    \includegraphics[width=\linewidth]{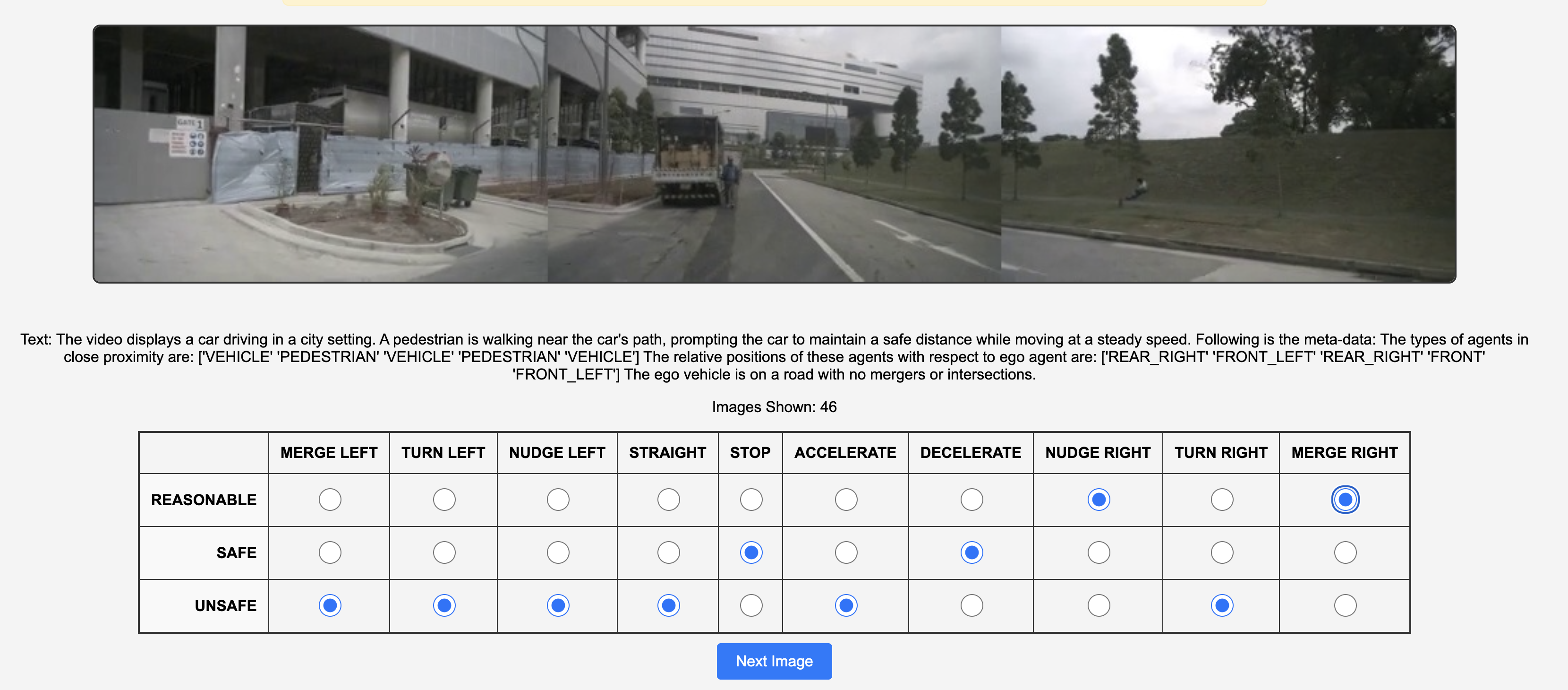}
    \caption{The custom human labelling interface. Labellers are shown a driving scene image and text description, then asked to classify each of the 10 possible ego vehicle actions as SAFE, UNSAFE, or REASONABLE based on the current driving context.}
    \label{fig:labeller-interface}
\end{figure}

The labelling process ensures that each action is evaluated in the context of the specific driving scenario, allowing labellers to consider factors such as traffic conditions, road geometry, presence of other vehicles, and overall driving context when making their safety assessments. This human-annotated dataset serves as the ground truth for evaluating our systems.
\end{document}